%% file: main.tex
\documentclass{article} 
\usepackage{iclr2019_conference,times}

\input{math_commands.tex}

\usepackage{hyperref}
\usepackage{url}

\usepackage{wrapfig}
\usepackage{graphicx}
\usepackage{subcaption}
\usepackage{amsmath,amsthm, amssymb}
\newtheorem{thm}{Theorem}
\newcommand*{\QED}{\hfill\ensuremath{\blacksquare}}
\usepackage[algoruled,noend]{algorithm2e}

\title{Learning Programmatically Structured \\ Representations with Perceptor Gradients}

\author{
  Svetlin Penkov \textsuperscript{\normalfont 1, 2, }\thanks{Work done as part
    of the author's PhD thesis at the University of Edinburgh.}\hspace{2mm}, Subramanian Ramamoorthy \textsuperscript{\normalfont 1, 2}\\
  \textsuperscript{1}The University of Edinburgh, \textsuperscript{2}FiveAI \\
  \texttt{\{sv.penkov,  s.ramamoorthy\}@ed.ac.uk}\\
}

\iclrfinalcopy 
\begin{document}

\maketitle

\begin{abstract}
  We present the perceptor gradients algorithm -- a novel approach to learning
  symbolic representations based on the idea of decomposing an agent's policy
  into i) a perceptor network extracting symbols from raw observation data and
  ii) a task encoding program which maps the input symbols to output actions. We
  show that the proposed algorithm is able to learn representations that can be
  directly fed into a Linear-Quadratic Regulator (LQR) or a general purpose A*
  planner. Our experimental results confirm that the perceptor gradients
  algorithm is able to efficiently learn transferable symbolic representations
  as well as generate new observations according to a semantically meaningful
  specification.
\end{abstract}

\section{Introduction}

Learning representations that are useful for devising autonomous agent policies
to act, from raw data, remains a major challenge. Despite the long history of
work in this area, fully satisfactory solutions are yet to be found for
situations where the representation is required to address symbols whose
specific meaning might be crucial for subsequent planning and control policies
to be effective. So, on the one hand, purely data-driven neural network based
models, e.g., \citep{duan2016benchmarking, arulkumaran2017deep}, make
significant data demands and also tend to overfit to the observed data
distribution \citep{srivastava2014dropout}. On the other hand, symbolic methods
that offer powerful abstraction capabilities, e.g., \citep{kansky2017schema,
  verma2018}, are not yet able to learn from raw noisy data. Bridging this gap
between neural and symbolic approaches is an area of active research
\citep{garnelo2016, garcez2018, keser17, penkov17icml}.

In many applications of interest, the agent is able to obtain a coarse `sketch'
of the solution quite independently of the detailed representations required to
execute control actions. For instance, a human user might be able to provide a
programmatic task description in terms of abstract symbols much easier than more
detailed labelling of large corpora of data. In the literature, this idea is
known as end-user programming \citep{lieberman2006enduser}, programming by
demonstration \citep{billard2008robot} or program synthesis
\citep{gulwani2017program}. In this space, we explore the specific hypothesis
that a programmatic task description provides inductive bias that enables
significantly more efficient learning of symbolic representations from raw data.
The symbolic representation carries semantic content that can be grounded to
objects, relations or, in general, any pattern of interest in the environment.

We address this problem by introducing the \textit{perceptor gradients}
  algorithm which decomposes a typical policy, mapping from observations to
actions, into i) a \textit{perceptor} network that maps observations to symbolic
representations and ii) a user-provided task encoding \textit{program} which is
executed on the perceived symbols in order to generate an action. We consider
both feedforward and autoencoding perceptors and view the program as a
regulariser on the latent space which not only provides a strong inductive bias
structuring the latent space, but also attaches a semantic meaning to the learnt
representations. We show that the perceptor network can be trained using the
REINFROCE estimator \citep{williams1992simple} for \textit{any} task encoding
program. 

We apply the perceptor gradients algorithm to the problem of balancing a
cart-pole system with a Linear-Quadratic Regulator (LQR) from pixel
observations, showing that the state variables over which a concise control law
could be defined is learned from data. Then, we demonstrate the use of the
algorithm in a navigation and search task in a Minecraft-like environment,
working with a 2.5D rendering of the world, showing that symbols for use within
a generic A* planner can be learned. We demonstrate that the proposed algorithm
is not only able to efficiently learn transferable symbolic representations,
but also enables the generation of new observations according to a semantically
meaningful specification. We examine the learnt representations and show that
programmatic regularisation is a general technique for imposing an inductive
bias on the learning procedure with capabilities beyond the statistical
constraints typically used.






\section{Related Work}




Representation learning approaches rely predominantly on imposing statistical
constraints on the latent space \citep{bengio2013representation} such as
minimising predictability \citep{schmidhuber1992learning}, maximising
independence \citep{barlow1989finding, higgins2016}, minimising total
correlation \citep{chen2018, kim2018} or imposing structured priors on the
latent space \citep{chen2016, siddarth2017}.
While learning disentangled features is an important problem, it does not by
itself produce features of direct relevance to the planning/control task at
hand. For example, the `best' features describing a dynamic system may be
naturally entangled, as in a set of differential equations. A programmatic task
representation allows such dependencies to be expressed, making subsequent
learning more efficient and improving generalisation capabilities
\citep{kansky2017schema, kusner2017grammar, gaunt2017, verma2018}.

Indeed, the idea of leveraging domain knowledge and imposing model driven
constraints on the latent space, has been studied in different domains. Much of
this work is focused on learning representations for predicting and controlling
physical systems where various assumptions about the underlying dynamic model
are made \citep{watter2015embed, iten2018discovering, fraccaro2017, karl2017}.
\citet{bezenac2018} even leverage a theoretical fluid transport model to
forecasting sea surface temperature. Model based representation learning
approaches have also been applied to inverse graphics, where a visual renderer
guides the learning process \citep{mansinghka2013approximate,
  kulkarni2014inverse, ellis2017learning}, and inverse physics, where a physics
simulator is utilised \citep{wu2017learning}. Interestingly,
\citet{kulkarni2015picture} propose a renderer that takes a programmatic
description of the scene as input, similar to the one used in
\citet{ellis2017learning}. Of particular relevance to our work is the idea of
learning compositional visual concepts by enforcing the support of set theoretic
operations such as union and intersection in the latent space
\citep{higgins2017scan}. Programmatic regularisation, which we propose, can be
viewed as a general tool for expressing such model based constraints.

Programmatic task descriptions can be obtained through natural language
instructions \citep{kaplan2017, Matuszek2013}, demonstrations in virtual
environments \citep{penkov17icml} or even manually specified. Importantly,
problems that appear quite hard from a pixel-level view can actually be more
easily solved by constructing simple programs which utilise symbols that exploit
the structure of the given problem. Sometimes, utilising prior knowledge in this
way has been avoided in a puristic approach to representation and policy
learning. In this paper, we address the more pragmatic view wherein the use of
coarse task descriptions enables the autonomous agents to efficiently learn
more abstract representations with semantic content that are relevant to practice.




\section{Problem Definition}

\begin{figure}
  \centering
  \includegraphics[scale=0.45]{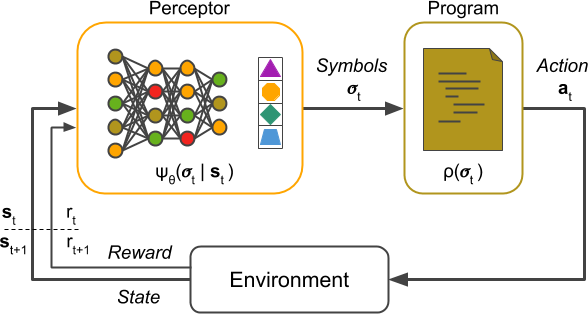}
  \vspace{0.2cm}
  \caption{The proposed policy decomposition into a perceptor and a program.}
  \label{fig:perceptor-setup}
\end{figure}

Let us consider the Markov decision process (MDP) represented as the tuple $(\mathcal{S},
\mathcal{A}, P, r, \gamma, P_0)$ where $\mathcal{S}$ is the set of possible
states, $\mathcal{A}$ is the set of possible actions, $P$ is the state
transition probability distribution, $r: \mathcal{S} \times \mathcal{A}
\rightarrow \mathbb{R}$ is the reward function, $\gamma$ is the reward
discounting factor and $P_0$ is the probability distribution over the initial
states. At each time step $t$, the agent observes a state $\mathbf{s}_t \in
\mathcal{S}$ and chooses an action $\mathbf{a}_t \in \mathcal{A}$ which results
in a certain reward $r_t$. We consider the stochastic policy
$\pi_{\theta}(\mathbf{a}_t | \mathbf{s}_t)$ belonging to a family of functions
parameterised by $\theta$ such that $\pi_{\theta}: \mathcal{S} \rightarrow
\mathcal{P}_{\mathcal{A}}\left(\mathcal{A} \right)$ where
$\mathcal{P}_{\mathcal{A}}$ is a probability measure over the set $\mathcal{A}$.

We are interested in decomposing the policy $\pi_{\theta}$ into i) a perceptor
$\psi_{\theta}: \mathcal{S} \rightarrow \mathcal{P}_{\Sigma}(\Sigma)$, where
$\Sigma$ is a set of task related symbols and $\mathcal{P}_{\Sigma}$ is a
probability measure over it, and ii) a task encoding program $\rho: \Sigma
\rightarrow \mathcal{A}$ such that $\rho \circ \psi_{\theta} : \mathcal{S}
\rightarrow \mathcal{A}$ as shown in Figure \ref{fig:perceptor-setup}. In this
paper, we address the problem of learning $\psi_{\theta}$ with any program $\rho$
and investigate the properties of the learnt set of symbols $\Sigma$.

\section{Method}

Starting from an initial state of the environment $\mathbf{s}_0 \sim
P_0(\mathbf{s})$ and following the policy $\pi_{\theta}$ for $T$ time steps
results in a trace $\tau = (\mathbf{s}_0, \mathbf{a}_0, \mathbf{s}_1,
\mathbf{a}_1, \ldots, \mathbf{s}_T, \mathbf{a}_T)$ of $T + 1$ state-action pairs
and $T$ reward values $(r_1, r_2, \ldots, r_T)$ where $r_t = r(\mathbf{s}_t,
\mathbf{a}_t)$. The standard reinforcement learning objective is to find an
optimal policy $\pi_\theta^*$ by searching over the parameter space, such that
the total expected return
\begin{equation}
  J(\theta) = E_{\tau \sim p(\tau; \theta)}\left[ R_0(\tau) \right] = \int p(\tau; \theta) R_0(\tau) \;d\tau
\end{equation}
is maximised, where $R_t(\tau) = \sum_{i = t}^T{\gamma^{i - t} r(\mathbf{s}_t,
  \mathbf{a}_t)}$ is the return value of state $\mathbf{s}_t$.

The REINFORCE estimator \citep{williams1992simple} approximates the gradient of
the cost as
\begin{equation}
  \nabla_\theta J(\theta) \approx \frac{1}{n}\sum_{i=1}^{n}{\sum_{t=0}^{T-1}\nabla_{\theta}\log{\pi_\theta(\mathbf{a}_t^{(i)} | \mathbf{s}_t^{(i)})} \left( R_t(\tau^{(i)}) - b_{\phi}(\mathbf{s}_t^{(i)}) \right)}
  \label{eq:full-nstep-ac}
\end{equation}
where $n$ is the number of observed traces, $b_{\phi}$ is a baseline function
paramterised by $\phi$ introduced to reduce the variance of the estimator.

Let us now consider the following factorisation of the policy
\begin{equation}
  \pi_{\theta}(\mathbf{a}_t | \mathbf{s}_t) = p(\mathbf{a}_t | \bm{\sigma}_t) \psi_{\theta}(\bm{\sigma}_t | \mathbf{s}_t)
\end{equation}
where $\bm{\sigma}_t \in \Sigma$ are the symbols extracted from the state
$\mathbf{s}_t$ by the perceptor $\psi_\theta$, resulting in the augmented trace
$\tau = (\mathbf{s}_0, \bm{\sigma}_0, \mathbf{a}_0, \mathbf{s}_1, \bm{\sigma}_1, \mathbf{a}_1, \ldots,
\mathbf{s}_T, \bm{\sigma}_T, \mathbf{a}_T)$. We are interested in exploiting the programmatic
structure supported by the symbols $\bm{\sigma}_t$ and so we set
\begin{equation}
  p(\mathbf{a}_t | \bm{\sigma}_t) = \delta_{\rho(\bm{\sigma}_t)}(\mathbf{a}_t)
\end{equation}
which is a Dirac delta distribution centered on the output of the task encoding
program $\rho(\bm{\sigma}_t)$ for the input symbols $\bm{\sigma}_t$. Even though
the program should produce a single action, it could internally work with
distributions and simply sample its output. Decomposing the policy into a
program and a perceptor enables the description of programmatically structured
policies, while being able to learn the required symbolic representation from
data. In order to learn the parameters of the perceptor $\psi_{\theta}$, we
prove the following theorem.

\begin{thm}[Perceptor Gradients]\label{thm:perceptor_gradient}
For any decomposition of a policy $\pi_{\theta}$ into a program $\rho$ and a
perceptor $\psi_\theta$ such that
\begin{equation}
  \pi_{\theta}(\mathbf{a}_t | \mathbf{s}_t) = \delta_{\rho(\bm{\sigma}_t)}(\mathbf{a}_t) \; \psi_{\theta}(\bm{\sigma}_t | \mathbf{s}_t)
  \label{eq:policy_decomp}
\end{equation}
the gradient of the log-likelihood of a trace sample $\tau^{(i)}$ obtained by
following $\pi_{\theta}$ is
\begin{equation}
  \nabla_\theta \log{p(\tau^{(i)}; \theta)} = \sum_{t=0}^{T-1}\nabla_{\theta}\log{\psi_\theta(\bm{\sigma}_t^{(i)} | \mathbf{s}_t^{(i)})}
\end{equation}
\end{thm}
\textbf{Proof.} Substituting $\pi_{\theta}(\mathbf{a}_t | \mathbf{s}_t)$
according to (\ref{eq:policy_decomp}) gives
\begin{align}
  \nabla_\theta \log{p(\tau^{(i)}; \theta)}
  &= \sum_{t=0}^{T-1}\nabla_{\theta}\log{\pi_\theta(\mathbf{a}_t^{(i)} | \mathbf{s}_t^{(i)})} = \nonumber\\
  &= \sum_{t=0}^{T-1}\left[
    \nabla_{\theta}\log{\delta_{\rho(\bm{\sigma}_t^{(i)})}(\mathbf{a}_t^{(i)})} +
    \nabla_{\theta}\log{\psi_{\theta}(\bm{\sigma}_t^{(i)} | \mathbf{s}_t^{(i)})} \right]  = \nonumber\\
  &= \sum_{t=0}^{T-1} \nabla_{\theta}\log{\psi_{\theta}(\bm{\sigma}_t^{(i)} | \mathbf{s}_t^{(i)})} \nonumber
\hskip0.45\textwidth \QED
\end{align}
Theorem \ref{thm:perceptor_gradient} has an important consequence -- no matter
what program $\rho$ we choose, as long as it outputs an action in a finite
amount of time, the parameters $\theta$ of the perceptor $\psi_{\theta}$ can be
learnt with the standard REINFORCE estimator.

\vspace{-12pt}
\paragraph{Feedforward Perceptor}
By combining theorem \ref{thm:perceptor_gradient} with (\ref{eq:full-nstep-ac}), we
derive the following loss function
\begin{align}
  \mathcal{L}(\theta, \phi) &= \frac{1}{n}\sum_{i=1}^{n} \left\{ \mathcal{L}_{\psi}(\tau^{(i)}, \theta) + \mathcal{L}_{b}(\tau^{(i)}, \phi) \right\}
  \label{eq:feedforward-loss} \\
  \mathcal{L}_{\psi}(\tau^{(i)}, \theta) &= \sum_{t=0}^{T-1}\log{\psi_\theta(\bm{\sigma}_t^{(i)} | \mathbf{s}_t^{(i)})} \left( R_t(\tau^{(i)}) - b_{\phi}(\mathbf{s}_t^{(i)}) \right) \\
  \mathcal{L}_{b}(\tau^{(i)}, \phi) &= \sum_{t=0}^{T-1}\left( R_t(\tau^{(i)}) - b_{\phi}(\mathbf{s}_t^{(i)}) \right)^2
\end{align}
and $\tau^{(i)} = (\mathbf{s}_0, \bm{\sigma}_0, \mathbf{s}_1, \bm{\sigma}_1,
\ldots, \mathbf{s}_T, \bm{\sigma}_T)$ contains $T + 1$ sampled state-symbol
pairs.
Algorithm \ref{alg:perceptor_rollout} demonstrates how to rollout a
policy decomposed into a perceptor and a program in order to obtain trajectory
samples $\tau^{(i)}$, while the overall perceptor gradients learning procedure
\setlength{\intextsep}{-20pt}%
\begin{wrapfigure}{R}{0.49\textwidth}
\input{pg/algo-feedforward-rollout}
\input{pg/algo-feedforward-pg}
\end{wrapfigure}
is summarised in Algorithm \ref{alg:perceptor_gradients}.

\paragraph{Autoencoding Perceptor}
The perceptor is a mapping such that  $\psi_{\theta}: \mathcal{S} \rightarrow
\mathcal{P}_{\Sigma}(\Sigma)$ and so by learning the inverse mapping
$\omega_{\upsilon}: \Sigma \rightarrow \mathcal{P}_{\mathcal{S}}(\mathcal{S})$,
parameterised by $\upsilon$, we enable the generation of states (observations)
from a structured symbolic description. Thus $\psi_\theta$ and $\omega_\upsilon$
form an autoencoder, where the latent space is $\Sigma$. Importantly, the
resulting autoencoder can be trained efficiently by applying the
reparameterisation trick \citep{kingma2013, jang2016categorical} when sampling
values for $\bm{\sigma}_t$ during the perceptor rollout and reusing the obtained
samples for the training of the generator $\omega_\upsilon$. In order to do so,
we simply augment the loss in (\ref{eq:feedforward-loss}) with a reconstruction
loss term
\begin{align}
  \mathcal{L}_{\omega}(\tau^{(i)}, \theta) &= \sum_{t=0}^{T-1}\log{\omega_\upsilon(\mathbf{s}_t^{(i)} | \bm{\sigma}_t^{(i)})}
\end{align}

\section{Experimental Results}
\subsection{Cart-Pole Balancing}
\begin{figure}
  \centering
  \includegraphics[scale=0.39]{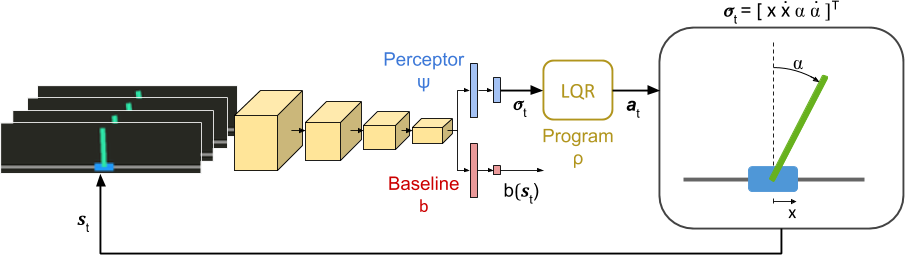}
  \caption{A diagram of the cart-pole experimental setup.}
  \label{fig:cart-pole-model}
\end{figure}

We first consider the problem of balancing a cart-pole system by learning
symbolic representations from the raw image observations. The cart-pole
system is well studied in optimal control theory and it is typically balanced
with an LQR \citep{zhou1996robust}. We exploit this knowledge and set the
program $\rho$ to implement an LQR. The perceptor $\psi_\theta$ is a
convolutional neural network (see \ref{sec:A1}) as shown in the overall
experiment diagram in Figure \ref{fig:cart-pole-model}. We define the state
vector as
\begin{equation}
  \bm{\sigma} =
  \begin{bmatrix}
    x & \dot{x} & \alpha & \dot{\alpha}
  \end{bmatrix}^T
\end{equation}
where $x \in \mathbb{R}$ is the linear position of the cart and $\alpha \in
\mathbb{R}$ is the angle of the pendulum with respect to its vertical position
as shown in Figure \ref{fig:cart-pole-model}.

\paragraph{LQR Program}
Given the standard state space representation of a cart-pole system (see
\ref{sec:A3}) we can design a feedback control law $u = -\mK \bm{\sigma}$, where
$u$ corresponds to the force applied to the cart and $\mK$ is a gain matrix. In
an LQR design, the gain matrix $\mK$ is found by minimising the quadratic cost
function
\begin{equation}
  J = \int_0^{\infty}\bm{\sigma}(t)^T \mQ \bm{\sigma}(t) + u(t)^T \mR u(t)\;dt
  \label{eq:lqr-loss}
\end{equation}
where we set $\mQ = 10^3\;\mI_4 $ and $R = 1$ as proposed in
\citep{lam2004control}. We set the program $\rho$ to be
\begin{equation}
  \rho(\bm{\sigma}) = \mathbf{a} =
  \begin{cases}
    1 & \quad \text{if } - \mK\bm{\sigma} > 0 \\
    0 & \quad \text{otherwise}
  \end{cases}
\end{equation}
producing 2 discrete actions required by the OpenAI gym cart-pole environment.
We used the \texttt{python-control}\footnote{\url{http://python-control.org}}
package to estimate ${\mK = [-1\; -2.25\;-30.74\; -7.07]}$.

\subsubsection{Learning Performance}
\begin{figure}
  \centering
  \includegraphics[scale=0.39]{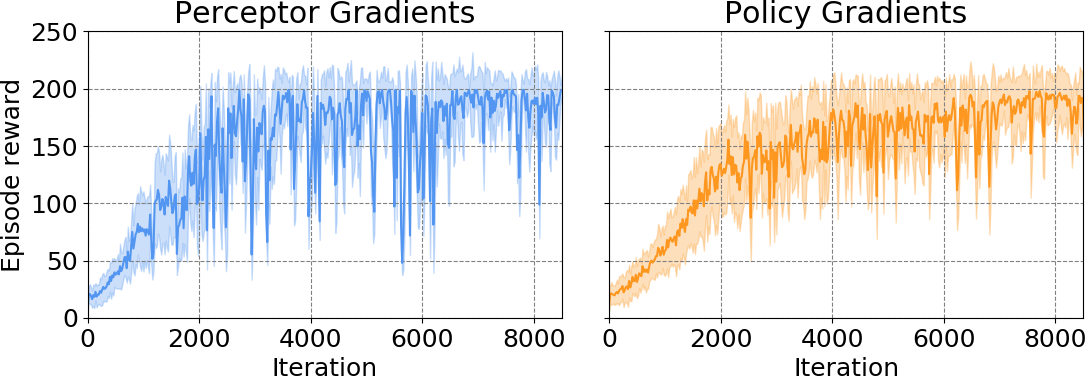}
  \caption{Learning performance at the cart-pole balancing task of the perceptor
    gradients algorithm (left) compared to
  standard policy gradients (right).}
  \label{fig:cart-pole-performance}
\end{figure}

In this experimental setup the perceptor is able to learn from raw image
observations the symbolic representations required by the LQR controller.
The average reward obtained during training is shown in Figure
\ref{fig:cart-pole-performance}. We compare the performance of the perceptor
gradients algorithm to a standard policy gradients algorithm, where we have
replaced the program with a single linear layer with sigmoid activation. The
perceptor obtains an average reward close to the maximum of 199 approximately
after 3000 iterations compared to 7000 iterations for the standard policy,
however the obtained reward has greater variance. Intuitively this can be
explained with the fact that the program encodes a significant amount of
knowledge about the task which speeds up the learning, but also defines a much
more constrained manifold for the latent space that is harder to be followed
during stochastic gradient descent.

\subsubsection{Perceptor Latent Space}
\begin{figure}
  \centering
  \includegraphics[scale=0.39]{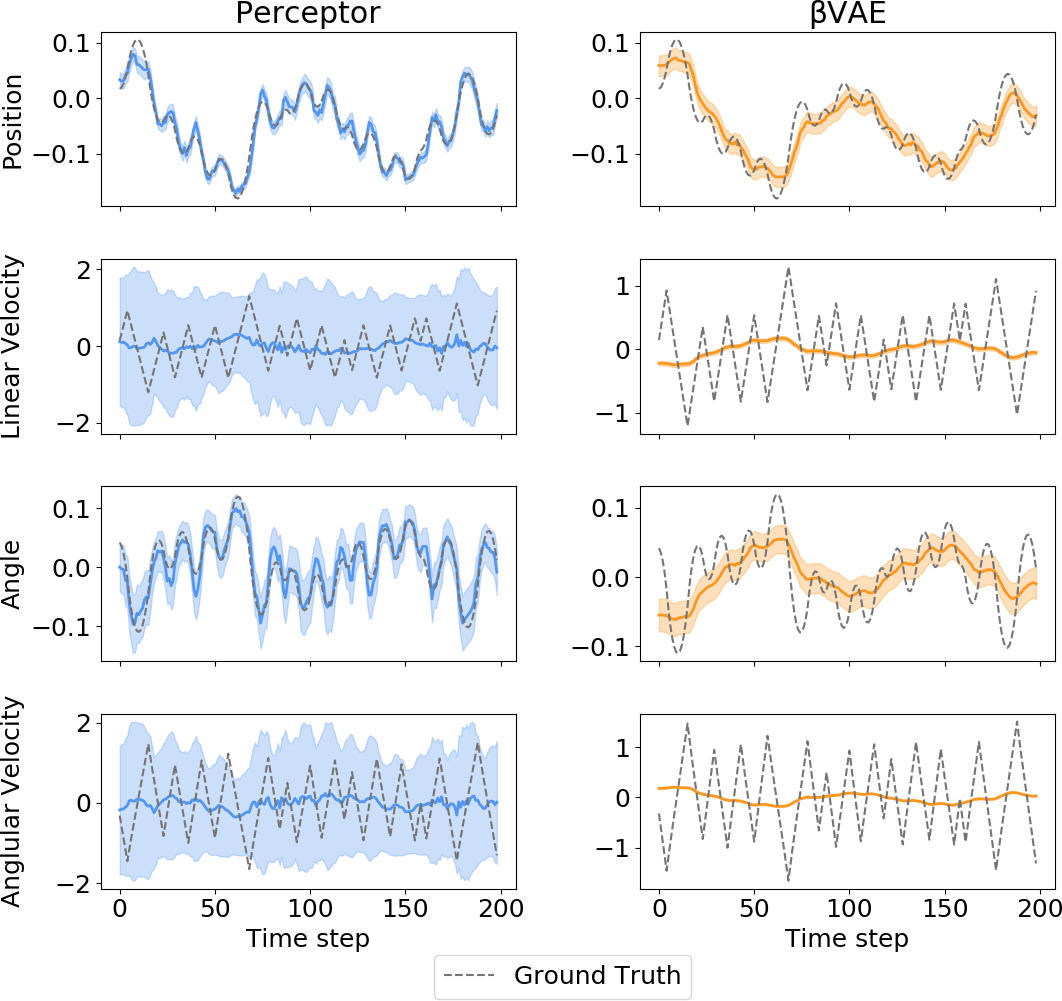}
  \caption{The latent space learnt by a perceptor (left) and $\beta$VAE (right).
  The thick coloured lines represent the predicted mean, while the shaded
  regions represent $\pm 1 \sigma$ of the predicted variance.}
  \label{fig:cart-pole-latent-space}
\end{figure}

The state space for which the minimum of the LQR cost in (\ref{eq:lqr-loss}) is
obtained is defined only up to scale and rotation (see \ref{sec:A4}). Therefore, we
use one episode of labelled data to find the underlying linear transformation
through constrained optimisation. The transformed output of the perceptor for an
entire episode is shown in Figure \ref{fig:cart-pole-latent-space}. The position
and angle representations learnt by the perceptor match very closely the ground
truth. However, both the linear and angular velocities do not match the true
values and the associated variance is also quite large. Investigating the
results, we found out that the LQR controller is able to balance the system even
if the velocities are not taken into account. Given that the performance of the
task is not sensitive to the velocity estimates, the perceptor was not able to
ground the velocities from the symbolic state representation to the
corresponding patterns in the observed images.

We compared the representations learnt by the perceptor to the ones learnt by a
$\beta$VAE \citep{higgins2016}. In order to do so, we generated a dataset of
observations obtained by controlling the cart-pole with the perceptor for 100
episodes. We set the $\beta$VAE encoder to be architecturally the same as the
perceptor and replaced the convolutional layers with transposed convolutions for
the decoder. We performed linear regression between the latent space of the
trained $\beta$VAE and the ground truth values. As it can be seen from the
results in Figure \ref{fig:cart-pole-latent-space} the representations learnt by
the $\beta$VAE do capture some of the symbolic state structure, however they are
considerably less precise than the ones learnt by the perceptor. The $\beta$VAE
does not manage to extract the velocities either and also collapses the
distribution resulting in high certainty of the wrong estimates.

As proposed by \citet{higgins2016}, setting $\beta > 1$ encourages the
$\beta$VAE to learn disentangled representations. The results shown in Figure
\ref{fig:cart-pole-latent-space}, though, were obtained with $\beta = 0.1$,
otherwise the regularisation on the latent space forcing the latent dimensions
to be independent was too strong. It has been recognised by the community that
there are better ways to enforce disentanglement such as minimising total
correlation \citep{chen2018, kim2018}. However, these methods also rely on the
independence assumption which in our case does not hold. The position and the
angle of the cart-pole system are clearly entangled and the program is able to
correctly bias the learning procedure as it captures the underlying
dependencies.

\subsection{Minecraft: Go to Pose}

\begin{figure}
  \centering
  \includegraphics[scale=0.39]{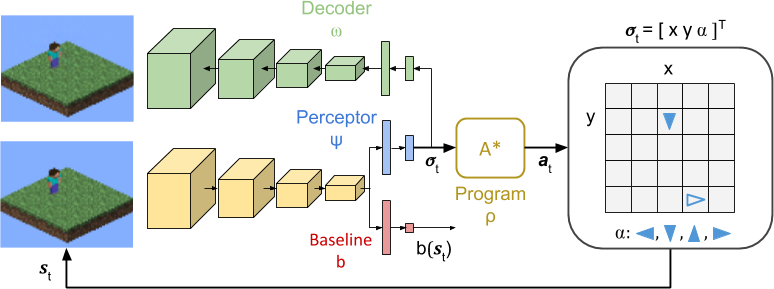}
  \caption{A diagram of the 'go to pose' experimental setup.}
  \label{fig:goto-loc-model}
\end{figure}

\paragraph{Task Description}
We apply the perceptor gradients algorithm to the problem of navigating to a
certain pose in the environment by learning symbolic representations from
images. In particular, we consider a $5 \times 5$ grid world where an agent is
to navigate from its location to a randomly chosen goal pose. The agent receives
+1 reward if it gets closer to the selected goal pose, +5 if it reaches the
position and +5 if it rotates to the right orientation. To encourage optimal paths,
the agent also receives -0.5 reward at every timestep. A symbolic representation
of the task together with the 2.5D rendering of the environment and the
autoencoding perceptor, which we train (see \ref{sec:A2}), are shown in Figure
\ref{fig:goto-loc-model}. We express the state of the environment as
\begin{equation}
  \bm{\sigma} =
  \begin{bmatrix}
    x & y & \alpha
  \end{bmatrix}^T
\end{equation}
where $x, y \in \{1, 2, 3, 4, 5\}$ are categorical variables representing the
position of the agent in the world and $\alpha \in \{1, 2, 3, 4\}$ represents its
orientation.

\paragraph{A* Program}
Given that the pose of the agent $\bm{\sigma}$ and the goal pose $G$ are known
this problem can be easily solved using a general purpose planner. Therefore, in
this experiment the program $\rho$ implements a general purpose A* planner. In
comparison to the simple control law program we used in the cart-pole
experiments, $\rho$ in this case is a much more complex program as it contains
several loops, multiple conditional statements as well as a priority queue. For
the experiments in this section we directly plugged in the implementation
provided by the \texttt{python-astar} package
\footnote{\url{https://pypi.org/project/astar/}}.

At every timestep, a path is found between the current $\bm{\sigma}_t$ produced
by the perceptor and the goal $G$ randomly chosen at the beginning of the
episode such that the agent either moves to one of the 4 neighbouring squares or
rotates in-place at $90^{\circ}$, $180^\circ$ or $270^\circ$. The output action
$\mathbf{a}_t = \rho(\bm{\sigma_t})$ is set to the first action in the path
found by the A* planner.

\subsubsection{Learning Performance}

\begin{figure}
  \centering
  \includegraphics[scale=0.39]{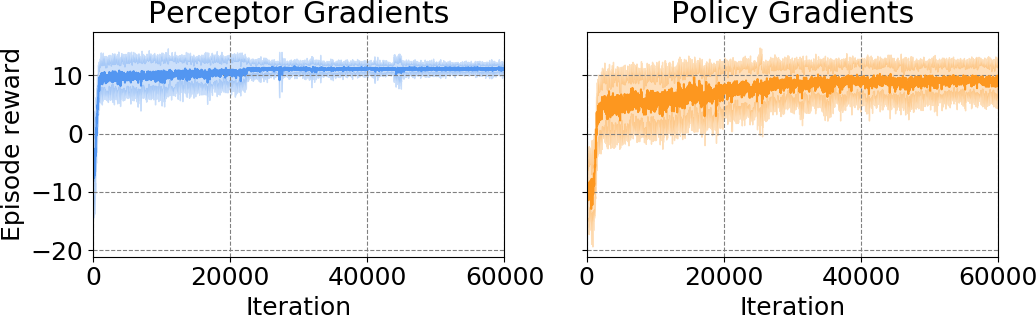}
  \caption{Learning performance at the 'go to pose' task of the perceptor gradients (left)
    compared to the policy gradients (right).}
  \label{fig:goto-loc-performance}
\end{figure}

In this experimental setup the perceptor is able to learn the symbolic
representations required by the A* planner from raw image observations.
The average reward obtained during training is shown in Figure
\ref{fig:goto-loc-performance}. Again, we compare the performance of the
perceptor gradients to a standard policy gradients algorithm, where we have replaced the
program with a single linear layer with softmax output. Additionally, we
rendered an arrow in the observed images to represent the chosen goal such that
the policy network has access to this information as well. In only 2000
iterations, the perceptor obtains an average reward close to the optimal one of
approximately $11.35$, while it takes more than 30000 iterations for the policy
gradients algorithm to approach an average reward of 10. Furthermore, given the
highly structured nature of the environment and the task, the perceptor
gradients agent eventually learns to solve the task with much greater
reliability than the pure policy gradients one. See \ref{sec:A5} for
experimental results on the Minecraft tasks with a feedforward perceptor instead
of an autoencoding one.

\subsubsection{Perceptor Latent Space}
\begin{figure}
  \centering
  \includegraphics[scale=0.30]{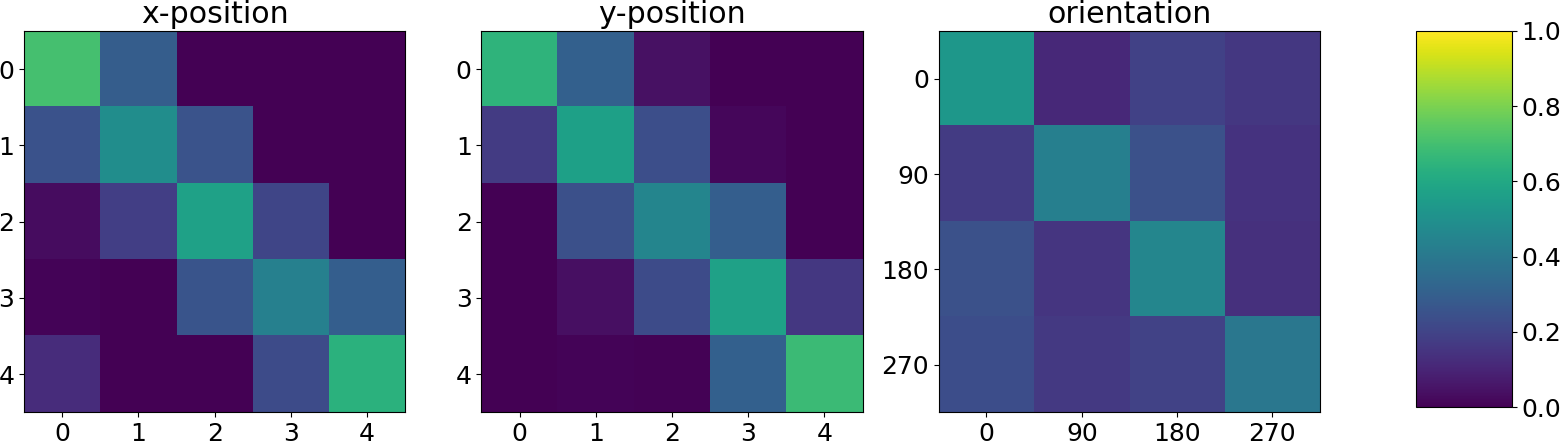}
  \caption{Confusion matrices between the values predicted by the perceptor
    (horizontally) and the true values (vertically) for each of the symbolic
    state components.}
  \label{fig:goto-loc-confusion-matrix}
\end{figure}

In this case the latent space directly maps to the symbolic state description of
the environment and so we collected a dataset of 1000 episodes and compared the
output of the perceptor to the ground truth values. The results, summarised
in the confusion matrices in Figure \ref{fig:goto-loc-confusion-matrix},
indicate that the perceptor has learnt to correctly ground the symbolic state
description to the observed data. Despite the fact that there are some errors,
especially for the orientation, the agent still learnt to perform the task
reliably. This is the case because individual errors at a certain timestep have
a limited impact on the eventual solution of the task due to the robust nature
of the abstractions supported by the symbolic A* planner.

\subsubsection{Generating Observations}
\begin{figure}
  \centering
  \includegraphics[scale=0.39]{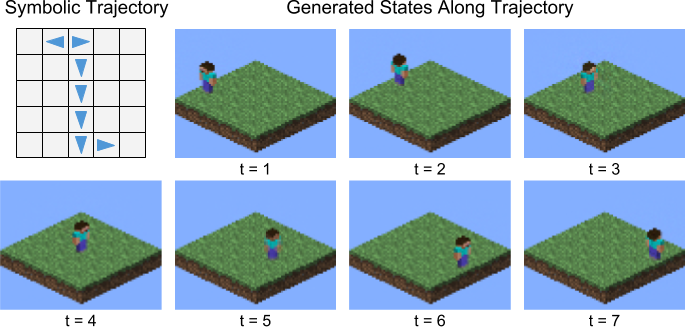}
  \caption{Sampled images of states following a symbolically defined
    trajectory.}
  \label{fig:goto-loc-decoded}
\end{figure}
Since we have trained an autoencoding perceptor we are also able to generate new
observations, in this case images of the environment. Typically, manual
inspection of the latent space is needed in order to assign each dimension to a
meaningful quantity such as position or orientation. However, this is not the
case for autoencoding perceptors, as the program attaches a semantic meaning to
each of the symbols it consumes as an input. Therefore, we can generate
observations according to a semantically meaningful symbolic specification
without any inspection of the latent space. In Figure
\ref{fig:goto-loc-decoded}, we have chosen a sequence of symbolic state
descriptions, forming a trajectory, that we have passed through the decoder in
order to generate the corresponding images. Given the lack of noise in the
rendering process we are able to reconstruct images from their symbolic
descriptions almost perfectly. Nevertheless, some minor orientation related
artefacts can be seen in the generated image corresponding to $t=5$.

\subsection{Minecraft: Collect Wood}
\begin{figure}
  \centering
  \includegraphics[scale=0.39]{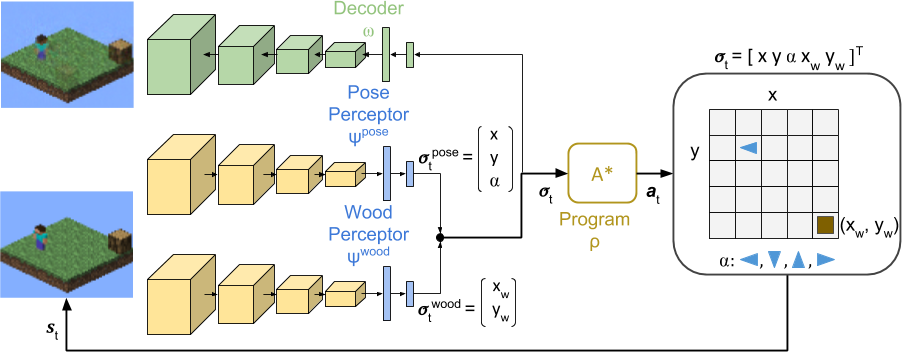}
  \caption{A diagram of the 'collect wood' experimental setup (baseline
    networks not illustrated).}
  \label{fig:pick-item-model}
\end{figure}

\paragraph{Task Description}
The last task we consider is navigating to and picking up an item, in particular
a block of wood, from the environment. In addition to the motion actions, the
agent can now also pick up an item if it is directly in front of it. The pick
action succeeds with 50\% chance, thus, introducing stochasticity in the task. As
discussed by \citet{levesque2005planning}, this problem cannot be solved with a
fixed length plan and requires a loop in order to guarantee successful
completion of the task. The agent receives +5 reward whenever it picks up a
block of wood. We expand the state to include the location of the wood block
resulting in
\begin{equation}
  \bm{\sigma} =
  \begin{bmatrix}
    x & y & \alpha & x_w & y_w
  \end{bmatrix}^T
\end{equation}
where $x_w, y_w \in \{1, 2, 3, 4, 5\}$ are categorical variables representing the
location of the block of wood and $x$, $y$, and $\alpha$ are defined as before.
The symbolic state representation as well as the rendered observations are shown
in Figure \ref{fig:pick-item-model}.

\paragraph{Stacked Perceptors}
In the cart-pole experiment the controller balances the system around a fixed
state, whereas in the navigation experiment the A* planner takes the agent to a
randomly chosen, but known, goal pose. Learning to recognise both the pose of the
agent and the position of the item is an ill posed problem since there is not a
fixed frame of reference -- the same sequence of actions can be successfully
applied to the same relative configuration of the initial state and the goal,
which can appear anywhere in the environment. We have, however, already trained
a perceptor to recognise the pose of the agent in the 'go to pose' task and so
in this experiment we demonstrate how it can be transferred to a new task. We
combine the pose perceptor from the previous experiment with another perceptor
that is to learn to recognise the position of the wood block. Given the symbolic
output of both perceptors, we can simply concatenate their outputs and feed them
directly into the A* program, as shown in Figure \ref{fig:pick-item-model}. Even
though the symbols of the pose perceptor are directly transferable to the
current task, it needs to be adapted to the presence of the unseen before wooden
block. We train both perceptors jointly, but keep the learning rate for the
pre-trained pose perceptor considerably lower than the one for the wood
perceptor. This ensures that no catastrophic forgetting occurs during the
initial stages of training.

\subsubsection{Learning Performance}
\begin{figure}
  \centering
  \includegraphics[scale=0.39]{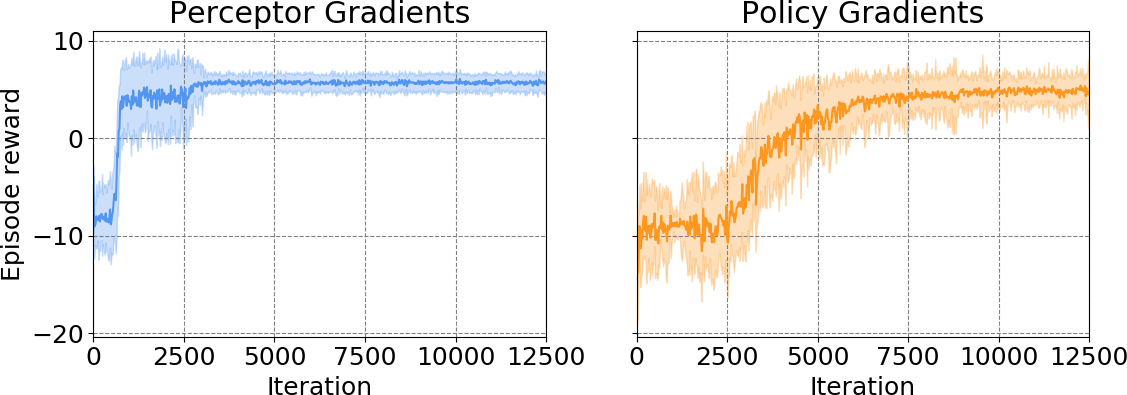}
  \caption{Learning performance at the 'collect wood' task of the perceptor gradients (left)
    compared to the policy gradients (right).}
  \label{fig:pick-item-performance}
\end{figure}
As shown in Figure \ref{fig:pick-item-performance} the agent is quickly able to
solve the task with the stacked configuration of perceptors. It significantly
outperforms the neural network policy that is trained on the entire problem and
achieves optimal performance in less than 3000 iterations. These result clearly
demonstrate that the symbolic representations learnt by a perceptor can be
transferred to new tasks. Furthermore, perceptors not only can be reused, but
also adapted to the new task in a lifelong learning fashion, which is
reminiscent of the findings in \citep{gaunt2017} and their idea of neural
libraries.

\subsubsection{Generating Observations}
\begin{figure}
  \centering
  \includegraphics[scale=0.39]{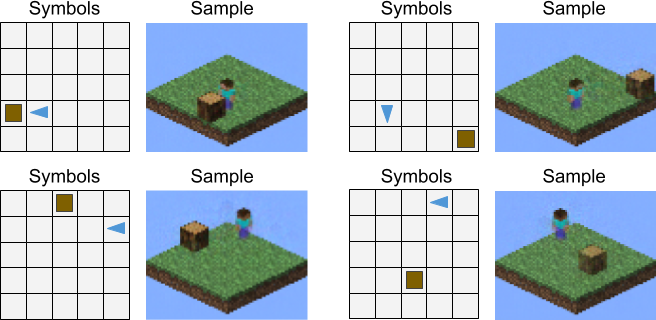}
  \caption{Sampled images from a symbolic specification over the joint latent
    space of the pose and wood perceptors.}
  \label{fig:pick-item-decoded}
\end{figure}
Stacking perceptors preserves the symbolic nature of the latent space and so
we are again able to generate observations from semantically meaningful
specifications. In Figure \ref{fig:pick-item-decoded} we have shown a set of
generated state samples from a symbolic specification over the joint latent
space. The samples not only look realistic, but also take occlusions into
account correctly (e.g. see top left sample).

\section{Conclusion}

In this paper we introduced the perceptor gradients algorithm for learning
programmatically structured representations. This is achieved by combining
a perceptor neural network with a task encoding program. Our approach achieves
faster learning rates compared to methods based solely on neural networks and
yields transferable task related symbolic representations which also carry
semantic content. Our results clearly demonstrate that programmatic
regularisation is a general technique for structured representation learning.


\bibliography{iclr2019_conference}
\bibliographystyle{iclr2019_conference}
\newpage
\appendix
\section{Supplementary Information}

\subsection{Cart-Pole Feedforward Perceptor}
\label{sec:A1}
\begin{figure}
  \centering
  \includegraphics[scale=0.35]{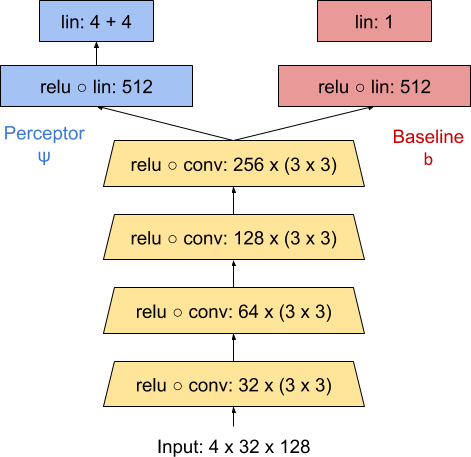}
  \caption{Architecture of the cart-pole feedforward perceptor $\psi_\theta$ and
    the baseline network $b_\phi$. Convolutions are represented as \#filters
    $\times$ filter size and all of them have a stride of 2. Linear layers are
    denoted by the number of output units.}
  \label{fig:cart-pole-perceptor}
\end{figure}

The input of the cart-pole feedforward perceptor is a stack of 4 consecutive
grayscale $32 \times 128$ images that we render the cart-pole system onto as
shown in Figure \ref{fig:cart-pole-model}. This is a setup similar to the one
proposed in \citep{mnih2015} which preserves temporary information in the input
such that it can be processed by a convolutional neural network. The
architecture of the perceptor $\psi_\theta$ is shown in Figure
\ref{fig:cart-pole-perceptor}. Note that the perceptor shares its convolutional
layers with the baseline network $b_\phi$. The outputs of the perceptor are the
mean and the diagonal covariance matrix of a 4-dimensional normal distribution.

\subsection{Minecraft 'Go-to Pose' Autoencoding Perceptor}
\label{sec:A2}
\begin{figure}
  \centering
  \includegraphics[scale=0.35]{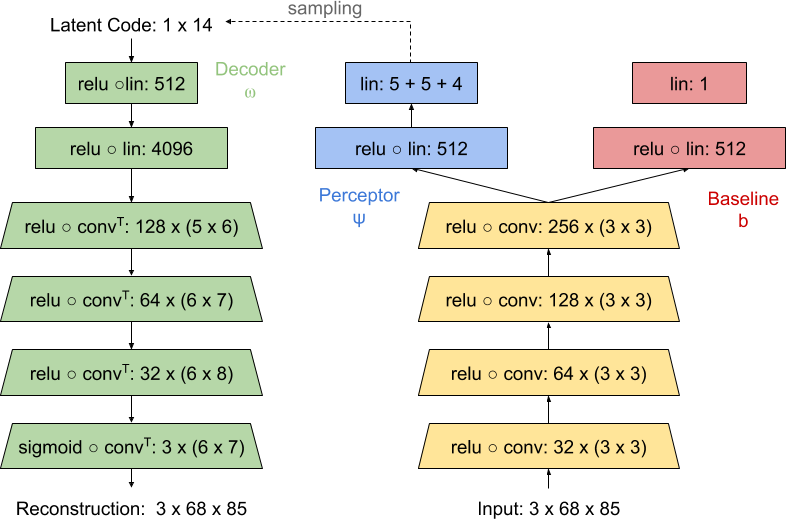}
  \caption{Architecture of the 'go-to pose' autoencoding perceptor
    $\psi_\theta$, the decoder $\omega_\upsilon$ and the baseline network
    $b_\phi$. Convolutions are represented as \#filters $\times$ filter size and
    all of them have a stride of 2. Transposed convolutions ($conv^T$) are
    represented in the same way. Linear layers are denoted by the number of
    output units.}
  \label{fig:goto-loc-perceptor}
\end{figure}

For this experiment we designed an autoencoding perceptor, the architecture of
which is shown in Figure \ref{fig:goto-loc-perceptor}. The input is a single
color image containing 2.5D rendering of the world as shown in Figure
\ref{fig:goto-loc-model}. The encoder outputs the parameters of 3 categorical
distributions corresponding to $x$, $y$ and $\alpha$ variables. These
distributions are sampled to generate a latent code that is put through the
decoder. We use the Gumbel-Softmax reparameterisation of the categorical
distributions \citep{jang2016categorical} such that gradients can flow from the
decoder through the latent code to the encoder.

\subsection{Cart-Pole State Space Model}
\label{sec:A3}
The state vector of the cart-pole system is
\begin{equation}
  \bm{\sigma} =
  \begin{bmatrix}
    x & \dot{x} & \alpha & \dot{\alpha}
  \end{bmatrix}^T
\end{equation}
where $x \in \mathbb{R}$ is the linear position of the cart and $\alpha \in
\mathbb{R}$ is the angle of the pendulum with respect to its vertical position
as shown in Figure \ref{fig:cart-pole-model}. By following the derivation in
\citep{lam2004control} of the linearised state space model of the system around
the unstable equilibrium $[0 \; 0 \; 0 \; 0]^T$ (we ignore the modelling of the
gearbox and the motor) we set the system matrix $\mA$ and input matrix $\mB$ to
\begin{equation}
  \mA =
  \begin{bmatrix}
    0 & 1 & 0 & 0 \\
    0 & 0 & -\frac{gml}{LM - m l} & 0 \\
    0 & 0 & 0 & 1 \\
    0 & 0 & \frac{g}{L - m l / M} & 0
  \end{bmatrix}
  \quad
  \mB =
  \begin{bmatrix}
    0 \\
    \frac{1}{M - ml / L} \\
    0 \\
    -\frac{1}{ML - ml}
  \end{bmatrix}
\end{equation}
where $m$ is the mass of the pole, $M$ is the mass of the pole and the cart, $l$
is half of the pendulum length, $g$ is the gravitational acceleration, $L$ is
set to $\frac{I + ml^2}{ml}$ and $I=\frac{ml^2}{12}$ is the moment of inertia of
the pendulum. We use the cart-pole system from OpenAI gym where all the
parameters are pre-specified and set to $m = 0.1$, $M = 1.0$ and $l = 0.5$.

\subsection{LQR State Space Uniqueness}
\label{sec:A4}
Given that $\mQ$ is a matrix of the form $\lambda \mI_4$ where $\lambda$ is a
scalar, then any rotation matrix $\mM$ applied on the latent vector
$\bm{\sigma}$ will have no impact on the cost in (\ref{eq:lqr-loss}) as
\begin{equation}
  (\mM\bm{\sigma})^T \mQ \mM\bm{\sigma} = \bm{\sigma}^T\mM^T \lambda \mI_4 \mM\bm{\sigma} = \bm{\sigma}^T \lambda \mI_4 \mM^T \mM \bm{\sigma} = \bm{\sigma}^T \lambda \mI_4 \bm{\sigma} = \bm{\sigma}^T \mQ \bm{\sigma}
\end{equation}
since rotation matrices are orthogonal. Additionally, scaling $\bm{\sigma}$ only
scales the cost function and so will not shift the locations of the optima.

\subsection{Minecraft Tasks with a Feedforward Perceptor}
\label{sec:A5}

\begin{figure}
  \centering
  \includegraphics[scale=0.39]{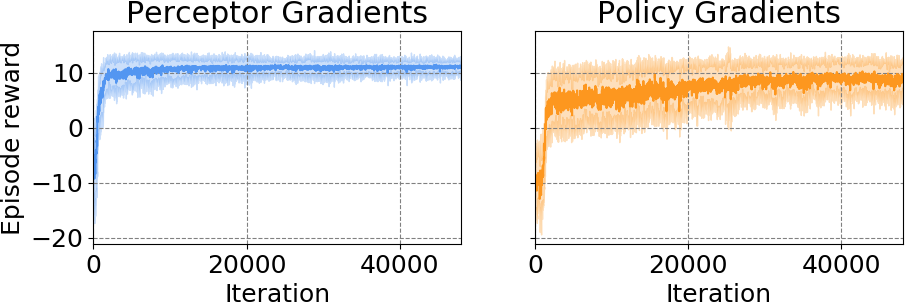}
  \caption{Learning performance with a feedforward perceptor at the 'go to
    pose' task of the perceptor gradients (left) compared to the policy
    gradients (right).}
  \label{fig:goto-loc-only-perceptor-performance}
\end{figure}

\begin{figure}
  \centering
  \includegraphics[scale=0.39]{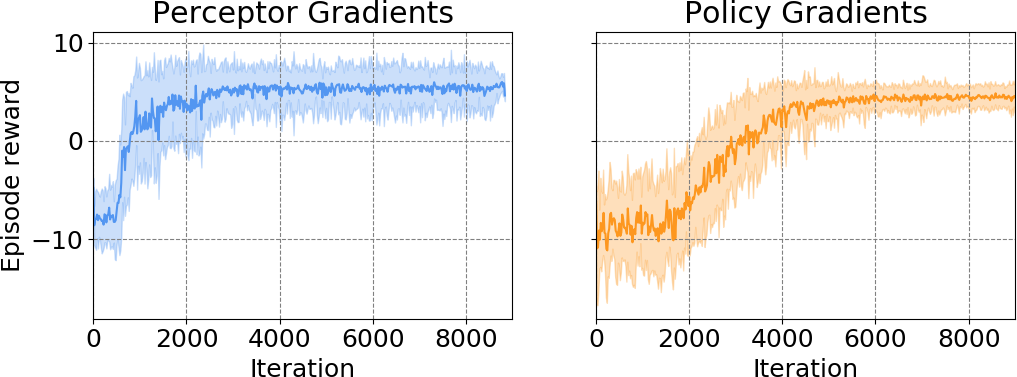}
  \caption{Learning performance with a feedforward perceptor at the 'collect
    wood' task of the perceptor gradients (left) compared to the policy
    gradients (right).}
  \label{fig:pick-item-only-perceptor-performance}
\end{figure}

In order to study the contribution of the decoder to the performance of the
agent in the Minecraft tasks we conducted a set of ablation experiments where we
replaced the autoencoding perceptor with a feedforward one. Figure
\ref{fig:goto-loc-only-perceptor-performance} and Figure
\ref{fig:pick-item-only-perceptor-performance} show the learning performance at
the 'go-to pose' and 'collect-wood' tasks, respectively, with a feedforward
perceptor. Overall, the results indicate that the main effect of the decoder is
to decrease the variance of the obtained reward during training. The feedforward
perceptor manages to ground the position of the agent slightly more accurately
than the autoencoding perceptor, however the accuracy of the orientation has
decreased. The reason for this is that orientation has little effect on the
performance of the agent as it can move to any square around it, regardless of
its heading. This is similar to the cart-pole task where the linear and angular
velocities had little effect on the LQR performance.

\begin{figure}
  \centering
  \includegraphics[scale=0.35]{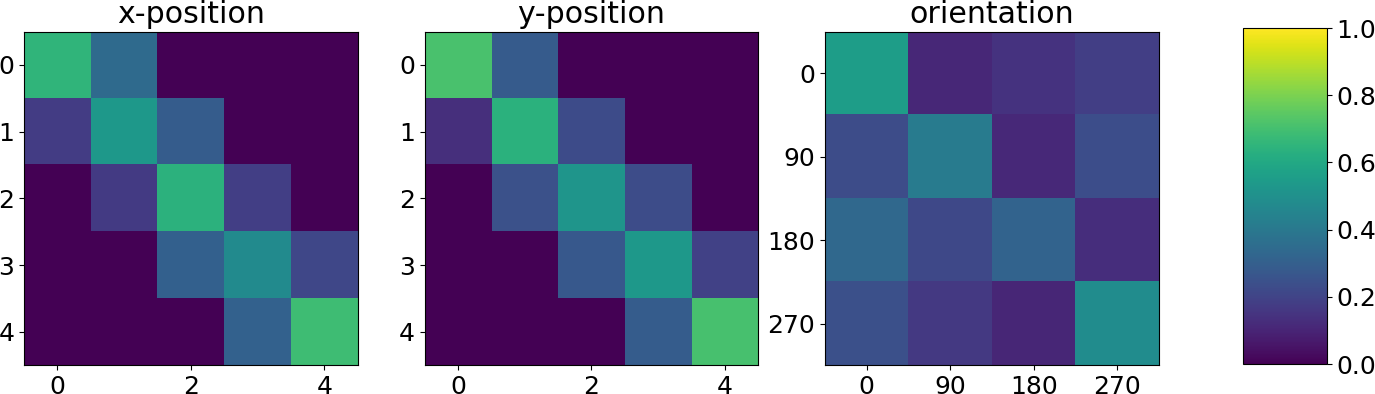}
  \caption{Confusion matrices between the values predicted by the feedforward perceptor
    (horizontally) and the true values (vertically) for each of the symbolic
    state components.}
  \label{fig:goto-loc-only-perceptor-confusion-matrix}
\end{figure}

\end{document}

%% file: math_commands.tex
\usepackage{amsmath,amsfonts,bm}

\def\eqref#1{equation~\ref{#1}}

\def\1{\bm{1}}

\def\mA{{\bm{A}}}
\def\mB{{\bm{B}}}

\def\mI{{\bm{I}}}

\def\mK{{\bm{K}}}

\def\mM{{\bm{M}}}

\def\mQ{{\bm{Q}}}
\def\mR{{\bm{R}}}

\DeclareMathAlphabet{\mathsfit}{\encodingdefault}{\sfdefault}{m}{sl}
\SetMathAlphabet{\mathsfit}{bold}{\encodingdefault}{\sfdefault}{bx}{n}



%% file: pg/algo-feedforward-rollout.tex
\begin{algorithm}[H]
	\caption{Perceptor rollout for a single episode}
    \label{alg:perceptor_rollout}
  	\KwIn{$\psi_\theta$, $\rho$}
    \KwOut{$\tau$, $r_{1:T}$} \BlankLine
    \For{$t=0$ \KwTo $T$} {
      $\mathbf{s}_t \leftarrow$ observe environment \\
      $\bm{\sigma}_t \leftarrow$ sample from $\psi_\theta(\bm{\sigma} | \mathbf{s}_t)$ \\
      $\mathbf{a}_t \leftarrow \rho(\bm{\sigma}_t)$ \\
      $r_t \leftarrow $ execute $\mathbf{a}_t$ \\
      append $(\mathbf{s}_t, \bm{\sigma}_t)$ to $\tau$
    }
\end{algorithm}

%% file: pg/algo-feedforward-pg.tex
\begin{algorithm}[H]
	\caption{Perceptor gradients}
    \label{alg:perceptor_gradients}
    $(\theta, \phi) \leftarrow$ Initialise parameters \\
    \Repeat{{\normalfont convergence of parameters} $(\theta, \phi)$} {
      \For{$i=1$ \KwTo $n$} {
        $\tau^{(i)}, r_{1:T}^{(i)} \leftarrow$ rollout$(\psi_\theta, \rho)$ \\
        \For{$t=0$ \KwTo $T$} {
          $R_t^{(i)} \leftarrow \sum_{i = t}^T{\gamma^{i - t} r_t^{(i)}}$ \\
          $A_t^{(i)} \leftarrow \left( R_t^{(i)} - b_{\phi}(\mathbf{s}_t^{(i)}) \right)$
        }
      } \BlankLine
      $\mathcal{L}_\psi \leftarrow \frac{1}{n}\sum_{i=1}^{n}
      \sum_{t=0}^{T-1}\log{\psi_\theta(\bm{\sigma}_t^{(i)} |
        \mathbf{s}_t^{(i)})} A_t^{(i)}$ \\
      $\mathcal{L}_b \leftarrow \frac{1}{n}\sum_{i=1}^{n} \sum_{t=0}^{T-1}
      \left( R_t^{(i)} - b_{\phi}(\mathbf{s}_t^{(i)}) \right)^2$ \BlankLine
      $\mathcal{L} \leftarrow \mathcal{L}_\psi + \mathcal{L}_b$ \\
      $\mathbf{g} \leftarrow \nabla_{\theta,\; \phi} \mathcal{L}(\theta, \phi)$ \BlankLine
      $(\theta, \phi) \leftarrow$ Update parameters using $\mathbf{g}$ \BlankLine
    }
\end{algorithm}